\documentclass[runningheads]{llncs}
\usepackage[T1]{fontenc}
\usepackage{graphicx}
\usepackage{textcomp}
\usepackage{array,ragged2e}
\usepackage{colortbl}
\usepackage{makecell}
\usepackage{pgfplots}
\pgfplotsset{compat=1.17}
\usepackage{amsmath}
\usepackage{multicol}
\usepackage{multirow}
\usepackage{tabto}
\usepackage{wrapfig}
\usepackage{amssymb}
\usepackage{amsmath}
\usepackage{float}
\usepackage{graphicx}
\usepackage[misc,geometry]{ifsym}
\usepackage{lipsum}
\usepackage[utf8]{inputenc}
\DeclareUnicodeCharacter{03BF}{o}

\usepackage{textcomp}
\usepackage{array,ragged2e}
\usepackage{pgfplots}
\usepackage{url}
\usepackage{lineno,hyperref}
\hypersetup{
  linkcolor  = red!60!black,
  citecolor  = blue!90!white,
  urlcolor   = violet!60!black,
  colorlinks = true,
}

\usepackage{orcidlink}

\usepackage{hyperref}
\usepackage{svg}
\usepackage{xcolor}
\newcommand{\orcid}[1]{\href{https://orcid.org/#1}{\includesvg[width=10pt]{orcid.svg}}}
\begin{document}
\title{Developing a Dyslexia Indicator Using Eye Tracking}
%
%
\title{Developing a Dyslexia Indicator Using Eye Tracking}

\author{Kevin Cogan\inst{1}\textsuperscript{$\ddagger$} \and
Vuong M. Ngo\orcidlink{0000-0002-8793-0504}\inst{2}~\textsuperscript{\Letter}\textsuperscript{$\ddagger$} \and
Mark Roantree\orcidlink{0000-0002-1329-2570}\inst{3}
}
\authorrunning{}

\institute{School of Computing, Dublin City University, Ireland \and
Ho Chi Minh City Open University, Ho Chi Minh City, Vietnam  \and
Insight Centre for Data Analytics, Dublin City University, Ireland
\email{kevin.s.cogan@gmail.com, vuong.nm@ou.edu.vn, mark.roantree@dcu.ie}
}

\maketitle              
\textsuperscript{$\ddagger$}{These authors designated as co-first authors.}

\begin{abstract}
Dyslexia, affecting an estimated 10\% to 20\% of the global population, significantly impairs learning capabilities, highlighting the need for innovative and accessible diagnostic methods. This paper investigates the effectiveness of eye-tracking technology combined with machine learning algorithms as a cost-effective alternative for early dyslexia detection. By analyzing general eye movement patterns, including prolonged fixation durations and erratic saccades, we proposed an enhanced solution for determining eye-tracking-based dyslexia features. A Random Forest Classifier was then employed to detect dyslexia, achieving an accuracy of 88.58\%. Additionally, hierarchical clustering methods were applied to identify varying severity levels of dyslexia. The analysis incorporates diverse methodologies across various populations and settings, demonstrating the potential of this technology to identify individuals with dyslexia, including those with borderline traits, through non-invasive means. Integrating eye-tracking with machine learning represents a significant advancement in the diagnostic process, offering a highly accurate and accessible method in clinical research.

\end{abstract}

\section{Introduction}

Dyslexia is commonly understood as a specific learning disability, characterized by difficulties with accurate and/or fluent word recognition, alongside poor spelling and decoding abilities which result in challenges with reading fluency, comprehension, and vocabulary growth \cite{gavin2016}. Neurologically, dyslexia is associated with differences in the brain's structure and function. Studies using functional magnetic resonance imaging (fMRI) and other neuroimaging techniques have identified atypical activations in regions of the brain responsible for key reading processes, including the left temporal-parietal cortex, which is involved in phonological processing and decoding \cite{bach2010}.

Dyslexia poses significant challenges in educational environments, impacting individuals' reading, writing, and language processing abilities. It is estimated to affect 10\% to 20\% of the global population \cite{wu2022}. Early detection is crucial for providing effective interventions, yet many face difficulties due to the high costs, limited access to specialized psychological assessments, and the heavy reliance on teachers to quickly identify learning difficulties.

Eye-tracking technology offers a non-invasive method to understand dyslexia by monitoring rapid eye movements, such as saccades and fixations. In the context of dyslexia, eye movement patterns differ significantly from those of non-dyslexic readers \cite{smyrnakis2021}. Individuals with dyslexia often exhibit longer fixation durations, more frequent and shorter saccades, and a higher number of regressions, reflecting difficulties in processing and decoding text. These distinctive eye movement patterns serve as biomarkers for dyslexia, allowing researchers and clinicians to explore the neurocognitive discrepancies that characterize the disorder. 

Dyslexia diagnosis has evolved from subjective methods to advanced technologies. The 1990s brought costly neuroimaging tools like fMRI and PET, revealing neurological differences \cite{richards2008}. By the 2010s, eye-tracking and machine learning emerged as efficient, non-invasive tools for analyzing reading behavior and identifying dyslexic traits \cite{nerusil2021}. Machine learning algorithms now enable the automatic classification of dyslexic and non-dyslexic readers based on eye movement data during reading \cite{poornappriya2020}, \cite{vajs2022b}, \cite{vajs2022}, \cite{vehlen2021}. This innovative approach leverages computational power to provide early, accurate, and efficient dyslexia diagnosis, facilitating timely interventions and support.

\textbf{Contribution.}
This paper investigates the use of eye-tracking technology as an innovative and cost-effective solution for dyslexia detection. By capturing precise eye movement metrics, such as saccades and fixations, eye-tracking offers a non-invasive and accessible method for identifying individuals with dyslexia. Building on foundational eye-tracking features from existing research, we propose an enhanced framework for developing eye-tracking-based dyslexia features. Machine learning techniques, including Random Forest Classifier and Agglomerative Hierarchical Clustering, are applied to analyze dyslexia-related patterns. The study evaluates the accuracy, effectiveness, and potential of eye-tracking in early dyslexia detection, highlighting its ability to complement traditional diagnostic methods. Additionally, it reviews research demonstrating the scalability and adaptability of eye-tracking across diverse populations and educational contexts, emphasizing its suitability for large-scale screening initiatives. By integrating advanced technologies, this paper aims to promote more inclusive, timely, and efficient dyslexia assessments.

\textbf{Paper Structure.}
The remainder of this paper is organized as follows: 
section \ref{sec:related_work} discusses related work; 
section \ref{sec:data_modeling}  outlines our data modeling approach and describes the solution for extracting enhanced eye-tracking-based dyslexia features, which are then used to apply machine learning algorithms; 
the experimental results and analyses for both classification and clustering tasks are presented in section \ref{sec:experiments}; 
and finally, section \ref{sec:conclusion} concludes the paper and suggests directions for future work.

\section{Related Work}
\label{sec:related_work}

The work presented in \cite{franzen2021} focuses on the comparison of reading strategies between individuals with dyslexia and non-dyslexic controls using eye movement analysis during standardized reading tasks. This study utilized the International Reading Speed Texts (IReST) for consistency in material complexity, with eye-tracking technology capturing critical metrics such as fixation duration, saccade length, and regression frequency. This approach sheds light on the visual sampling strategies employed by individuals with dyslexia.
    
Building on the application of eye-tracking, \cite{raatikainen2021} integrates machine learning techniques to diagnose developmental dyslexia, leveraging eye movement data from 165 participants in the eSeek project. Dyslexic individuals were identified based on low reading fluency scores, and the methodology employed Random Forest for feature selection and Support Vector Machines for classification. This dual approach effectively reduced the data’s dimensionality while addressing its complexity, underscoring the potential of machine learning for handling high-dimensional datasets in dyslexia research.

Similarly, \cite{vajs2022c} introduces the VGG16 model to analyze eye-tracking data for dyslexia detection in children aged 7 to 13. Unlike traditional preprocessing methods, this study transformed gaze data into color-coded 2D time series graphs, capturing intricate eye movement patterns. These graphs served as input for classification, with the study further exploring various network configurations and segmentation strategies to enhance the model’s accuracy.

Building on these advancements, Nerusil et al. \cite{nerusil2021} propose a groundbreaking method for dyslexia detection, utilizing Convolutional Neural Networks (CNNs) to process eye-tracking data holistically. The dataset comprises eye-tracking signals from 185 subjects, classified into low and high-risk categories for dyslexia. Departing from traditional methods, this approach minimizes preprocessing and eliminates explicit feature extraction, allowing the CNN to autonomously detect and classify dyslexia-related patterns directly from raw data. The signals are analyzed in both time and frequency domains, with data standardized through zero-padding or interpolation for time-domain representation and magnitude spectrum analysis for frequency-domain analysis.

Finally, \cite{smyrnakis2021} takes a feature-based machine learning approach, combining conventional eye-tracking metrics with novel features to characterize reading behaviors. The inclusion of innovative measures, such as the Fixation Intersection Coefficient (FIC) and Fixation Fractal Dimension (FFD), offers deeper insights into the spatial and temporal gaze dynamics associated with dyslexia. By employing various machine learning algorithms for classification, this study enriches our understanding of dyslexic reading patterns and their distinct characteristics.

In summary, the studies mentioned above did not incorporate a sufficient number of eye-tracking-based dyslexia features into their machine learning models, which likely limited the classification performance. Additionally, none of these projects focused on analyzing the correlations between features and clusters, missing an opportunity to extract more meaningful insights. Without this correlation analysis, valuable relationships between different eye-tracking metrics and dyslexia severity could not be fully leveraged to enhance the accuracy and interpretability of the models. We believe that by incorporating a broader set of features and performing correlation analyses, this offers the opportunity to significantly improve model performance and provide deeper insights into the underlying patterns of dyslexia.



\section{Feature Engineering}
\label{sec:data_modeling}
The Provo Corpus comprises quantitative eye-tracking data collected from participants at Brigham Young University \cite{luke2017}. The dataset, approximately 78MB in size, contains 230,412 entries spanning 63 columns of both numerical and categorical data. It includes detailed records of eye movements from proficient readers interacting with continuous text, along with predictability scores derived from the Predictability Norms surveys. These scores incorporate traditional cloze measures, as well as morpho-syntactic and semantic predictability metrics for each word. This dataset provides detailed insights into reading behaviors, such as fixations, saccades, and regressions. During the eye-tracking sessions, participants' eye movements were recorded using an SR Research EyeLink 1000 Plus eye-tracker, enabling precise measurement of their reading dynamics.

\subsection{Dataset Description}
\label{sec:dataset_desc}

\textbf{Predictability Norms Data.}
Predictability norms data were collected through a large-scale online survey involving 470 participants who were randomly assigned to complete five out of 55 texts. This helped quantify the predictability of each word within its context. The texts averaged 50 words and approximately 2.5 sentences each. In total, the dataset included 2,689 words, with 1,197 being unique word forms. Participants also completed an online survey via Qualtrics, where they provided demographic information and responded to text completion tasks. The texts were tagged for parts of speech using the Constituent Likelihood Automatic Word-Tagging System (CLAWS), and semantic association scores for each word were derived through latent semantic analysis (LSA).

\textbf{Eye-Tracking Data.} Eye-tracking data were gathered from 84 native English-speaking participants who read the same 55 texts used in the Predictability Norms group. The texts were sourced from a range of materials, including online news articles, popular science magazines, and public-domain fiction.

\subsection{Data Processing}
\textbf{Data cleaning:}  To streamline the model and improve its performance, redundant columns with high missing values, such as {\tt Ia\_Regression\_Path\_Duration} and {\tt Ia\_First\_Run\_Dwell\_Time}, were removed, along with highly correlated features like {\tt Ia\_First\_Saccade\_Start\_Time} and {\tt Ia\_First\_Saccade\_End\_Time}. Then, missing values were addressed using advanced techniques: the MissForest algorithm\footnote{\url{https://pypi.org/project/MissForest/}} was used to impute missing {\tt Ia\_Skip} values, rows lacking the {\tt Word} value were excluded, and sequential fields like {\tt Sentence\_Number} and {\tt Word\_In\_\-Sentence\_Number} were filled using the forward fill method. 
Integrating semi-structured data with categorical and continuous variables used the method presented in \cite{10.1093/comjnl/bxy064}. Finally, to manage outliers and ensure consistency, selected columns were normalized using MinMax Scaling, which adjusts the data to a specified range (0 to 1), enhancing the model's accuracy and robustness.

\textbf{Feature transformation:} We developed a new feature, {\tt Saccade\_Duration}, by calculating the time difference between {\tt Ia\_First\_Saccade\_End\_Time} and {\tt Ia\_First\_Saccade\_Start\_Time}, to more accurately measure the duration between saccades. Additionally, we transformed the {\tt Word\_Cleaned} column using the TF-IDF technique. This transformation enhances the analytical significance of each word by quantifying its importance relative to the document corpus, enabling more nuanced text analysis.

\vspace{-3mm}
\subsection{Enhanced Eye-Tracking-based Dyslexia Feature Determination}
\vspace{-1mm}
\subsubsection{a) Basic Dyslexia Features:}

Our research focuses on uncovering distinctions in eye-tracking metrics between individuals with and without dyslexia, building on insights from existing studies. By analyzing these differences through both statistical methods and visualizations—particularly by examining how these metrics vary across percentiles—we aim to identify meaningful patterns that can effectively distinguish skilled readers from those facing reading challenges. This dual approach not only highlights significant statistical disparities but also provides an intuitive understanding of the variations in reading behavior. By identifying how these metrics differ across readers, our study provides valuable insights into the cognitive and behavioral characteristics of dyslexia. Key metrics considered include:

\begin{itemize}
    \item {\tt Ia\_First\_Saccade\_Amplitude}: This metric quantifies the distance of the initial eye movement. Typically, dyslexic individuals exhibit shorter amplitudes, suggesting less efficient initial scanning of text \cite{franzen2021}.
    \item {\tt Ia\_Dwell\_Time}: This records the total time eyes fixate on a word. Extended dwell times in dyslexic readers imply greater effort in word recognition and processing \cite{franzen2021}.
    \item {\tt Ia\_Regression\_In\_Count} and {\tt Ia\_Regression\_Out\_Count}: \\
    These metrics count the instances where a reader's gaze regresses to previously read text, with dyslexics displaying more frequent regressions, a marker of disrupted reading fluency \cite{vagge2015}.
    \item {\tt Ia\_Fixation\_Count}: Captures the frequency of eye fixations on text, where higher counts are indicative of increased cognitive effort during reading, a common trait among dyslexics \cite{shalileh2023}.
    \item {\tt Saccade\_Duration}: Measures the duration of eye movements between fixations. Shorter durations in dyslexic readers reflect quick but inefficient scanning \cite{franzen2021}.
\end{itemize}

\textbf{Thresholds for labeling:} The 95th percentile was used as the threshold for each metric to differentiate between levels of reading difficulty. Any data entry above this threshold was assigned a label of 1, indicating a higher level of reading difficulty, while entries below were labeled 0, indicating lower difficulty. These labels were recorded in a new column named {\tt Reading\_Difficulties} \cite{raatikainen2021}.

\vspace{-3mm}
\subsubsection{b) Enhancing Dyslexia Features: } 
To determine the key factors influencing the \-{\tt Reading\_Difficulty} feature, a RandomForestClassifier was applied to a balanced dataset. The dataset was balanced by resampling the majority class to correct class imbalances, ensuring fairer model training and evaluation. Features directly contributing to the construction of the {\tt Reading\_Difficulty} target, such as {\tt Ia\_First\_Saccade\-\_Amplitude}, {\tt Ia\_Dwell\_Time}, and {\tt Ia\_Fixation\_\-Count}, were excluded from the analysis to avoid data leakage and ensure an unbiased assessment of feature importance.

The model was optimized using Bayesian hyperparameter tuning, which systematically explored the parameter space to maximize the classifier's performance. Once trained, the model identified the most influential features contributing to the prediction of reading difficulty. These key features were visualized using a bar plot, providing a clear, data-driven understanding of their relative importance and offering insights for refining the model further.

The most impactful features for predicting reading difficulty included: \\{\tt Ia\_Regression\_In\_Count}, {\tt Ia\_First\_Run\_Fixation\_.}, {\tt Saccade\_Duration},\\ {\tt Ia\_Right}, {\tt Ia\_First\_Fixation\_Time},
{\tt Ia\_Regression\_Out\_Count}, {\tt Ia\_First\_\-Fixation\_Index}, {\tt Ia\_First\_Fixation\_Y}, {\tt Ia\_First\_Fixation\_X}, {\tt Word\_Number}, {\tt Ia\_Skip} and {\tt Word\_Length}. These features, alongside the basic eye-tracking metrics described in Section 4a, form a comprehensive set of eye-tracking-enhanced dyslexia features. This enriched feature set aims to improve the detection and understanding of reading difficulties, particularly in dyslexic individuals. 

\subsection{Clustering and Classification Algorithms}

\subsubsection{a) Random Forest Classifier:}

Random Forest classification creates an ensemble of decision trees, each trained on a random subset of data through bootstrap sampling. At each node, a random subset of features is evaluated for splitting, promoting diversity and preventing any single feature from dominating the model. Each tree grows independently and makes its own prediction. For classification, the final prediction is based on majority voting across all trees \cite{CHOWDHURY2024}.

The hyperparameters of the Random Forest classifier is fine-tuned using BayesSearchCV, concentrating on the features selected during the forward selection process. This approach aims to optimize the cross-validation score by exploring various hyperparameter configurations. The Random Forest model is refined with key features identified through forward selection and benchmarked against a model trained on all features, with both employing optimized hyperparameters. To ensure robustness and manage variability, the model was trained on multiple resampled data subsets to address class imbalance.

\vspace{-3mm}
\subsubsection{b) Agglomerative Hierarchical Clustering:}
Agglomerative Hierarchical Clustering (AHC) is a bottom-up clustering method where each data point begins as its own cluster, and pairs of clusters are merged iteratively based on similarity until all data points belong to a single cluster or a predefined number of clusters are reached. We employ  Euclidean distance and Ward Linkage to calculate distances between clusters and determine how clusters are combined \cite{Ngo2021}.

One advantage of AHC is that it does not require the number of clusters to be specified beforehand, offering flexibility in exploratory data analysis. So, AHC is widely used in fields like bioinformatics, marketing, and text analysis due to its interpretability and ability to capture nested groupings of data points. Principal Component Analysis (PCA) \cite{KAMARI2022} is applied to reduce the dataset to two principal components. This reduction was essential for visualizing the clusters in a two-dimensional plot, providing immediate and clear visual insights into the complex patterns within the data.

\section{Evaluation and Discussion}
\label{sec:experiments}
In this section, we present a two-step validation process. First, we evaluate the performance of our dyslexia classifier, which identifies whether a reader likely displays dyslexia-related patterns. Second, we use cluster analysis to group readers by shared eye-tracking characteristics, offering deeper insight into the variability and potential severity of reading difficulties. Together, these complementary methods, classification for detection and clustering for subgroup analysis, deliver both strong diagnostic accuracy and a richer understanding of reading behavior, thereby reinforcing the robustness and interpretability of our overall findings.

\subsection{Classification Result}

The evaluation of the Random Forest Classifier was conducted using a 9-fold cross-validation process to ensure robustness and generalizability across various data subsets. This comprehensive evaluation aimed to establish a reliable baseline for the model's effectiveness in operational settings, with a focus on consistent performance metrics. Table \ref{tab:avg_kfold_metrics} presents the average metrics from the 9-fold cross-validation, reporting an accuracy of 88.58\%, precision of 87.91\%, recall of 89.49\%, and an F1-Score of 88.67\%.

\vspace{-3mm}
\begin{table}[H]
\centering
    \caption{Average 9-Fold Cross-Validation Metrics}
    \begin{tabular}{|l|c|c|c|c|}
    \hline
    \textbf{Metric} & \textbf{Accuracy (\%)} & \textbf{Precision (\%)} & \textbf{Recall (\%)} & \textbf{F1 Score (\%)}\\
    \hline
    \textbf{Average} & 88.58 & 87.91 & 89.49 & 88.67\\
    \hline
    \end{tabular}
    \label{tab:avg_kfold_metrics}
\end{table}
\vspace{-3mm}

\begin{figure}[H]
\small
    \begin{tabular}{ccc}
    \includegraphics[width=0.42\columnwidth]{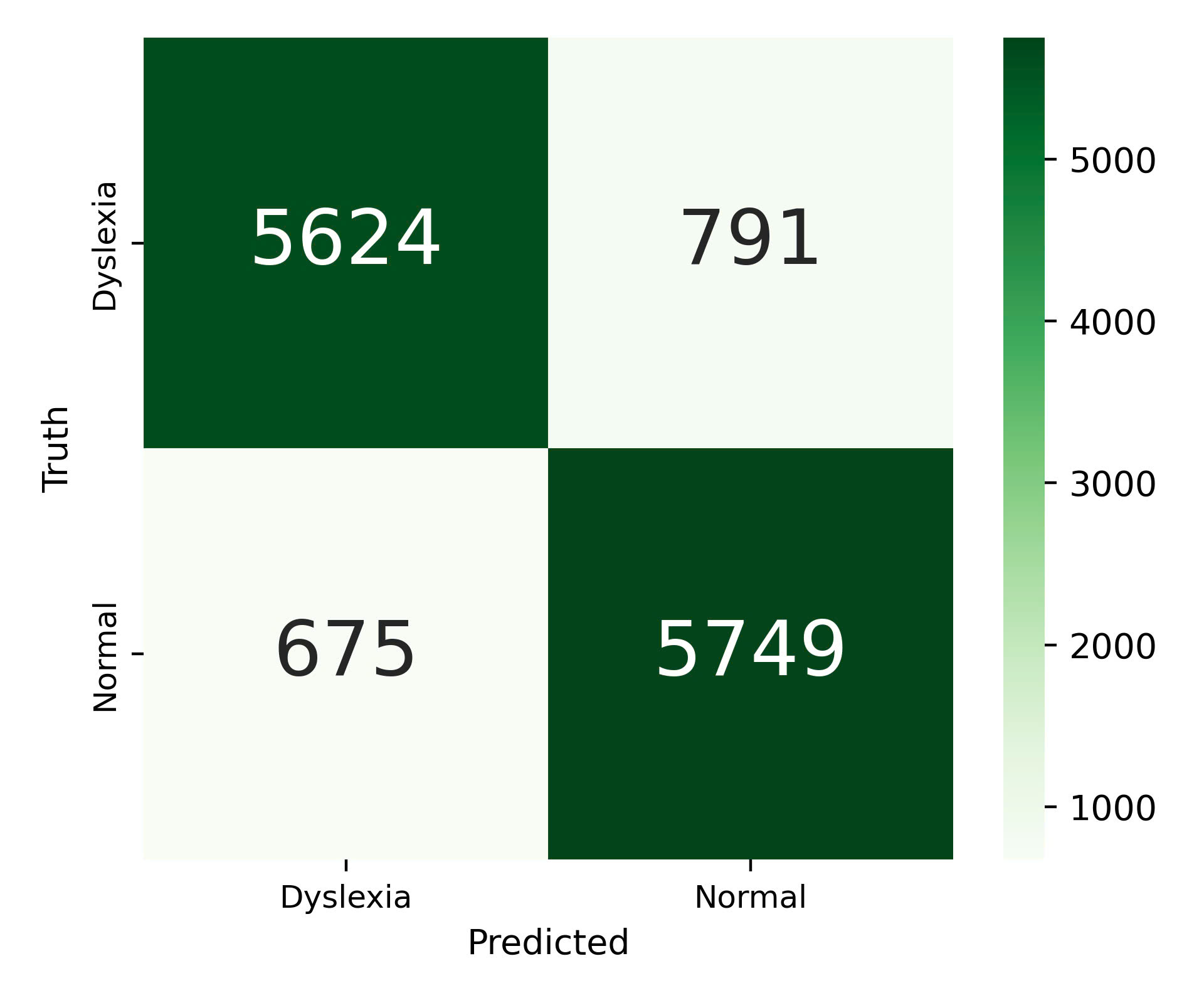} && \hspace{0mm}
    \includegraphics[width=0.5\columnwidth]{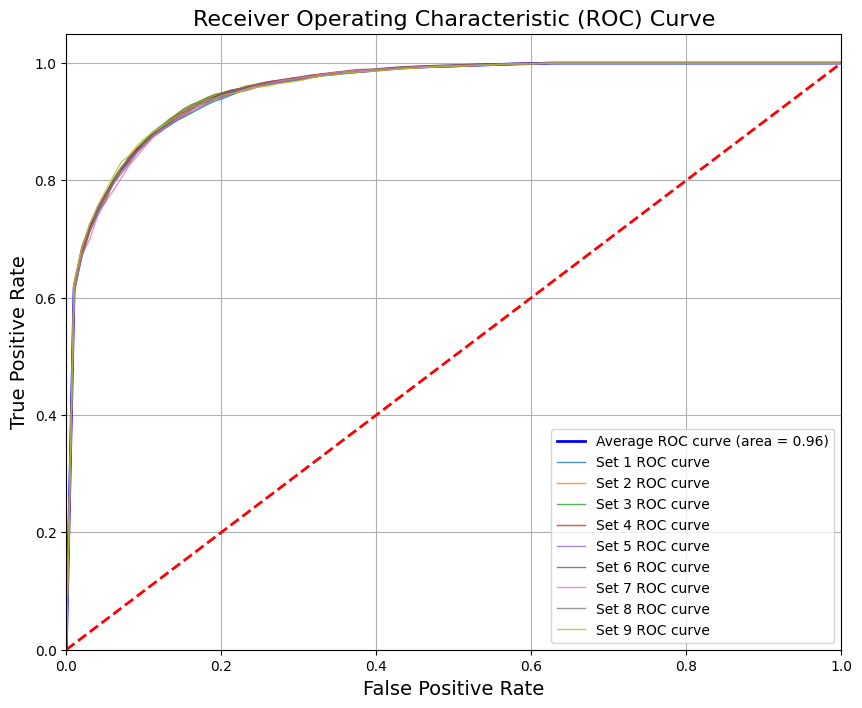}
    \vspace{2mm}
    \\
    (a) Average confusion matrix && \hspace{0mm} (b) ROC curves\\
    \end{tabular}
\vspace{-1mm}
\caption{Experimental Results on 9-Fold Cross-Validation}
\label{fig:expe-results}
\end{figure}

Additionally, Figure \ref{fig:expe-results}(a) displays the average confusion matrix, further highlighting the Random Forest Classifier's consistent performance across all validation folds. The model demonstrates a strong balance between sensitivity (high true positive rate) and specificity (high true negative rate), confirming its reliability and effectiveness in practical applications. This balance is particularly critical in scenarios where accurate decision-making is essential, and the cost of errors is high, making the model suitable for deployment in such environments.

To further evaluate the model’s performance, the Receiver Operating Characteristic (ROC) curve was plotted for each cross-validation fold, and the average ROC curve is presented below. This curve illustrates the trade-offs between true positive rates (sensitivity) and false positive rates for different threshold settings. Figure \ref{fig:expe-results}(b) displays the average Area Under the Curve (AUC) for the ROC curve, which is 0.96, indicating an excellent level of discrimination between the positive and negative classes. The consistency of the ROC curves across the nine folds suggests that the model's performance is robust and reliable across different data subsets. This high and consistent AUC value highlights the model's suitability for practical applications, such as diagnosing reading difficulties, where both high sensitivity and specificity are essential.

\vspace{-3mm}
\vspace{-5mm}

\vspace{-8mm}
\subsection{Cluster and Analysis}

\begin{figure}[H]
\centering
    \includegraphics[width=0.63\columnwidth]{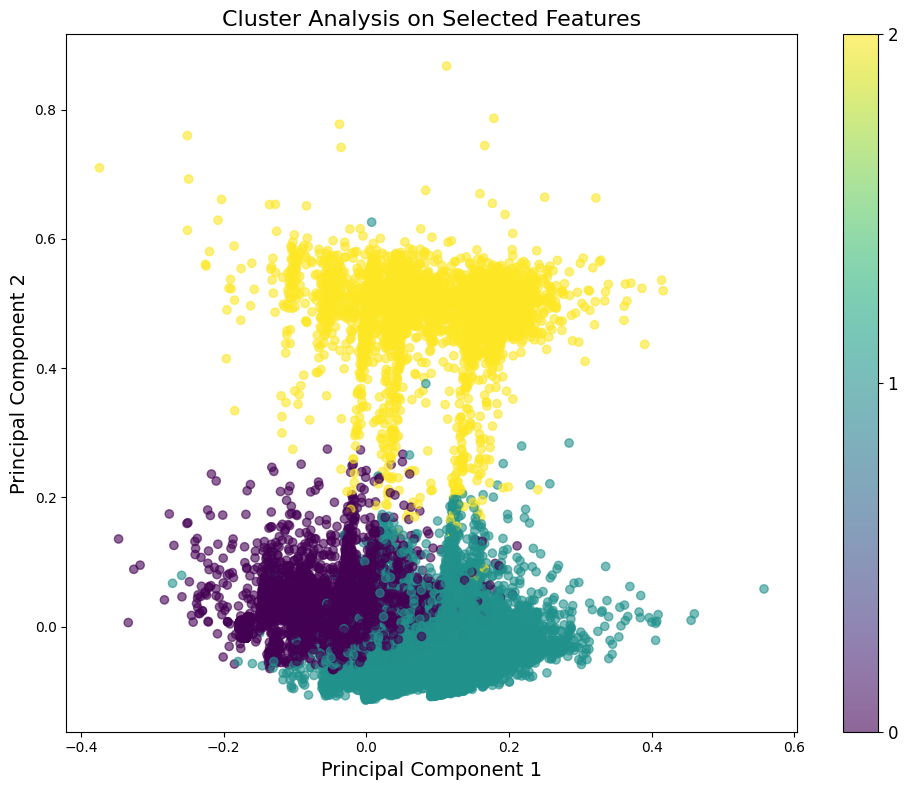}
    \vspace{-0.3cm}
    \caption{Cluster Analysis on Our Enhancing Dyslexia Features}
    \label{fig-cluster-enhanced-features}
\end{figure}

Figure \ref{fig-cluster-enhanced-features} presents a visualization of clustering results based on the enhanced eye-tracking-based dyslexia features, revealing three distinct clusters. Notably, there is a significant overlap between Clusters 0 and 1, indicating some shared characteristics between the groups. The detailed profiling of each cluster is as follows:

\begin{itemize}
\item Cluster 0: This cluster is characterized by low mean and median dwell times, shorter saccade durations, and moderate regression counts. These patterns suggest readers with efficient eye movement strategies, indicative of proficient readers who process text with minimal cognitive effort and high fluency.
\item  Cluster 1: Readers in this cluster exhibit higher dwell times, lower regression counts, and average saccade durations. These metrics reflect moderately proficient or average readers who may occasionally encounter difficulties but generally exhibit effective reading behaviors. This group likely includes readers with slightly reduced fluency compared to Cluster 0 but without significant impairments.
\item Cluster 2: This cluster is marked by significantly higher saccade amplitudes, increased dwell times, elevated fixation counts, and longer saccade durations. These patterns suggest a group of poor readers, potentially including individuals with dyslexia, who exhibit substantial effort and difficulty while reading. Their eye-tracking metrics indicate challenges in processing text efficiently, requiring more time and cognitive resources to decode words.
\end{itemize}

These findings provide a clear spectrum of reading behaviors, ranging from proficient to poor or dyslexic readers. The clustering results highlight the nuanced differences in eye movement patterns across the groups, with Cluster 2 particularly standing out for its association with reading difficulties. This visualization and clustering analysis underscore the potential of enhanced eye-tracking features to differentiate reading abilities. Such insights could inform the development of diagnostic tools for identifying dyslexia and tailoring targeted interventions to support individuals across the reading spectrum.

\vspace{-0.4cm}


\vspace{-1cm}
\section{Conclusion and Future Work}
\label{sec:conclusion}

The paper highlights the use of eye-tracking and machine learning to effectively detect dyslexia, leveraging metrics like saccade amplitude, dwell time, and fixation count. Operating at the 95th percentile, the method offers a non-invasive, objective, and cost-effective diagnostic tool, reducing reliance on subjective teacher judgment. This approach supports early identification of students needing intervention, enabling timely and targeted support \cite{gov2024}.

The development and evaluation of the Random Forest classifier underscored the pivotal roles of hyperparameter tuning and feature selection. Hyperparameter tuning optimized the model’s training process, enabling it to achieve superior results more efficiently. Selecting the top 12 essential features further enhanced performance while offering the potential to simplify future data collection and processing. The clustering analysis revealed distinct reading behaviors and strategies, illustrating a spectrum ranging from proficient readers to struggling ones, including individuals with dyslexic traits. These findings highlight the diversity in reading approaches and underscore the need for further research into targeted interventions and specialized support by identifying shared characteristics within each group.

Future research could focus on integrating eye-tracking data with other diagnostic methods, such as neuroimaging and behavioral assessments, to develop more comprehensive and multimodal approaches to dyslexia diagnosis. Combining these modalities would provide a richer understanding of the neural, cognitive, and behavioral mechanisms underlying dyslexia. For instance, neuroimaging could reveal brain activity patterns associated with reading difficulties, while behavioral assessments could capture additional contextual factors, such as phonological processing and working memory.

\vspace{-0.4cm}
\section*{Acknowledgment}
\vspace{-0.2cm}
This research has emanated from research conducted with the financial support of Taighde Éireann – Research Ireland under Grant number 12/RC/2289\_P2.

\vspace{-0.4cm}
\bibliographystyle{splncs04}
\bibliography{References.bib}

\end{document}